\documentclass[conference]{IEEEtran}
\IEEEoverridecommandlockouts
\usepackage{hyperref}
\usepackage{cite}
\usepackage{amsmath,amssymb,amsfonts}
\usepackage{algorithmic}
\usepackage{graphicx}
\usepackage{textcomp}
\usepackage[dvipsnames]{xcolor}
\def\BibTeX{{\rm B\kern-.05em{\sc i\kern-.025em b}\kern-.08em
    T\kern-.1667em\lower.7ex\hbox{E}\kern-.125emX}}
\usepackage{graphicx}
\usepackage{subcaption}
\graphicspath{ {images/} }
\usepackage[utf8]{inputenc}
\usepackage{fancyhdr}
\usepackage{dirtytalk}
\usepackage{tabularx}
 \usepackage{url}

\usepackage{url}

\usepackage[normalem]{ulem} 
\usepackage[ruled,vlined]{algorithm2e}

\usepackage{mathtools}

\DeclarePairedDelimiter\floor{\lfloor}{\rfloor}
\usepackage{enumitem,kantlipsum}
\usepackage{ulem}
\usepackage{caption}
\usepackage{xcolor}
\usepackage{svg}
\usepackage{graphicx}
\graphicspath{ {images/} }
\usepackage[utf8]{inputenc}
\usepackage{fancyhdr}
\usepackage{dirtytalk}
\usepackage{tikz}
\usetikzlibrary{shapes.geometric, arrows}
\usepackage{pgfplots}
\pgfplotsset{width=8cm,compat=1.16}
\usepgfplotslibrary{external}
\usepackage[nolist]{acronym}
\begin{acronym}
\acro{RAGAS}{RAG assessment}
\acro{LM}{language model}
\acro{LLM}{large language model}
\acro{SLM}{small language model}
\acro{RAG}{retrieval-augmented generation}
\acro{NLP}{natural language processing}
\acro{LoRA}{low-rank Adaptation}
\acro{QLoRA}{quantized low-rank adaptation}
\acro{SBERT}{sentence BERT}
\acro{ML}{machine learning}
\acro{OOD}{out-of-distribution}
\acro{QnA}{question and answer}
\acro{AI}{artificial intelligence}
\acro{PEFT}{parameter-efficient fine-tuning}

\acro{TeleQuAD}{telecom question answering dataset}
\acro{3GPP}{Third Generation Partnership Project}
\acro{O-RAN}{open radio access networks}
\acro{MCQ}{multiple choice question}

\acro{RoPE}{rotary position Embeddings}
\acro{TF-IDF}{term frequency inverse document frequency}

\end{acronym}

\tikzstyle{startstop} = [rectangle, rounded corners, minimum width=2cm, minimum height=0.5cm,text centered, draw=black, fill=red!30]
\tikzstyle{process} = [rectangle, minimum width=2cm, minimum height=0.5cm, text centered, draw=black, fill=orange!30, align=left]
\tikzstyle{decision} = [diamond, minimum width=1.0cm, minimum height=0.4cm, text centered, draw=black, fill=green!30]
\tikzstyle{arrow} = [thick,->,>=stealth]

\newcolumntype{L}{>{$}l<{$}}

\usepackage[labelformat=simple]{subcaption}

\begin{document}
\captionsetup[figure]{labelfont={bf},labelformat={default},labelsep=period,name={Figure}}

\title{TeleOracle: Fine-Tuned Retrieval-Augmented Generation with Long-Context Support for Networks}

\author{
Nouf~Alabbasi,~\IEEEmembership{Student~Member,~IEEE,}
         Omar~Erak,~\IEEEmembership{Student~Member,~IEEE,}      Omar~Alhussein,~\IEEEmembership{Member,~IEEE,}\\
         Ismail~Lotfi~\IEEEmembership{Member,~IEEE,}
         Sami Muhaidat,~\IEEEmembership{Senior~Member,~IEEE}
         Mérouane~Debbah,~\IEEEmembership{Fellow,~IEEE}
                    \thanks{This article was presented in part at the 2024 IEEE Global Communications Conference (Globecom).}
				\thanks{N. Alabbasi, O. Erak, O. Alhussein, I. Lotfi are with the KU 6G Research Center, Department of Computer Science, Khalifa University, Abu Dhabi, UAE (e-mail: Nouf.Alabbasi@ieee.org, omarerak@ieee.org, omar.alhussein@ku.ac.ae, ismail.lotfi@ku.ac.ae.)}
                    \thanks{S. Muhaidat is with the KU 6G Research Center, Department of Computer and Information Engineering, Khalifa University, Abu Dhabi, UAE, and with the Department of Systems and Computer Engineering, Carleton University, Ottawa, ON K1S 5B6, Canada, (e-mail: muhaidat@ieee.org).}
				\thanks{M. Debbah is with the KU 6G Research Center, Department of Computer and Information Engineering, Khalifa University, Abu Dhabi, UAE, , (e-mail: merouane.debbah@ku.ac.ae).}
		        
		\vspace{-0.7cm}		}

\maketitle

\begin{abstract}

The telecommunications industry's rapid evolution demands intelligent systems capable of managing complex networks and adapting to emerging technologies. While large language models (LLMs) show promise in addressing these challenges, their deployment in telecom environments faces significant constraints due to edge device limitations and inconsistent documentation.
To bridge this gap, we present TeleOracle, a telecom-specialized \ac{RAG} system built on the Phi-2 small language model (SLM). To improve context retrieval, TeleOracle employs a two-stage retriever that incorporates semantic chunking and hybrid keyword and semantic search. Additionally, we expand the context window during inference to enhance the model's performance on open-ended queries. We also employ low-rank adaption for efficient fine-tuning. A thorough analysis of the model's performance indicates that our \ac{RAG} framework is effective in aligning Phi-2 to the telecom domain in a downstream \ac{QnA} task, achieving a 30\% improvement in accuracy over the base Phi-2 model, reaching an overall accuracy of 81.20\%. Notably, we show that our model not only performs on par with the much larger LLMs but also achieves a higher faithfulness score, indicating higher adherence to the retrieved context.
\end{abstract}

\acresetall

    
\section{Introduction}    
The recent interest in \acp{LLM}, motivated by their unprecedented performance on various downstream tasks, has led to their widespread adoption across many domains. These models gained notable attention with a rising number of applications in text summarization, classification, and generation. In telecom, the potential gain from the incorporation of these models has not gone unnoticed \cite{largelanguagemodeldrivencurriculum,maatouk2024large, zhang2024interactive, lin2024primergenerativeaitelecom,Soman_2023,Zhou2024LargeLM}. The integration of \acp{LLM} into IoT devices represents a crucial step towards creating autonomous, agentic systems capable of making real-time decisions at the network edge~\cite{10439991}.

Foundation \ac{LLM} models, trained on vast amounts of diverse data, offer versatility and exhibit strong performance across generalized tasks. However, this broad knowledge can result in so-called negative transfer or knowledge interference in specialized domains such as telecommunications, where domain-specific terms and concepts often conflict with their general usage. Indeed, protocols and concepts in telecom sometimes follow distinct logic patterns that differ from broader contexts. This misalignment leads to poor performance when foundation \ac{LLM} models, such as GPT4, attempt to reconcile their general knowledge with telecom-specific concepts, as the former can actively interfere with the latter.
In addition, the strict latency requirements and resource constraints of edge devices, particularly in massive IoT deployments, hinder the deployment of LLMs, necessitating a shift toward smaller models~\cite{10722848}.

\Acp{SLM}, such as Microsoft's \texttt{Phi-2} with 2.7B parameters\cite{PHI2} and \texttt{Gemini Nano 2} with 3.2B parameters~\cite{gemini}, present a potential solution by offering competitive performance while maintaining the efficiency crucial for telecom applications.
Notably, despite its smaller size, Phi-2 performs on par with state-of-the-art \acp{LM} 25 times its size on various benchmarks. 
Telecom is a rapidly evolving domain that is continually adapting to accommodate the fast-paced technology development. As a result, technical papers and documents describing new ideas, standards, and protocols undergo frequent modifications, requiring \ac{AI} systems developed for the domain to be equally flexible in updating their knowledge.
This limits the utility of using fine-tuning to specialize the model to the domain. Unlearning in \acp{LLM} is also a significant challenge~\cite{yao2024largelanguagemodelunlearning}. Therefore, a more dynamic and adaptable approach, such as \ac{RAG} is needed for greater flexibility and cost-effectiveness. This approach enables 
the \acp{LM} to seamlessly incorporate new data without the need for extensive re-training. 

RAG minimizes hallucinations by grounding model responses in factual information through a two-step process: (1) storing domain-specific documents in a database (in semantic space), and (2) supplementing the user's query with relevant retrieved context at inference time.
The retrieved context can also include user-specific information or real-time system updates, allowing the model to leverage the user interaction history to produce tailored responses and remain aligned with current data. For these reasons, \ac{RAG} is particularly well suited for telecom applications.
However, while conventional \ac{RAG} models provide significant enhancement to language models' capabilities, telecom applications require special considerations due to the complex and technical nature of the domain. 
Telecom standardization documents do not follow a unified structure,
introducing problems when representing data in the \ac{RAG} database. This becomes an issue at the retrieval stage as the lack of standardization complicates the identification and extraction of relevant information, often resulting in suboptimal matches and reduced accuracy in responses. 
Additionally, common terms that recur across multiple areas in telecom documents often have nuanced meanings depending on the context. This contextual complexity necessitates careful
filtering of the retrieved documents to provide the generator with a list of highly relevant context chunks. These context chunks supplement the generator with the necessary information in order to allow it to adequately respond to the the given query. Moreover, open-ended or vague user queries require a larger amount of retrieved-context chunks. This demand can exceed the \ac{SLM}'s capacity, limiting its ability to handle the longer prompts required for such questions.
To address the aforementioned challenges, this paper makes the following contributions to specialized \acp{LM} for telecommunications:
\begin{itemize}
\item We develop TeleOracle, a specialized \ac{RAG} framework that effectively addresses the unique challenges of processing telecommunications documentation through strategic integration of multiple techniques;

\item We implement an optimized document processing pipeline that combines semantic chunking with a two-stage retrieval process, enabling precise and context-aware selection of relevant technical information;

\item We leverage SelfExtend to extend the generator's context window at inference time, accommodating a greater volume of retrieved information and enabling more comprehensive responses to complex or open-ended queries;

\item We conduct a comprehensive analysis of the proposed architecture. Notably, we show that our model not only performs on par with state-of-the-art \acp{LLM} models but also is higher on the faithfulness score, indicating higher adherence to retrieved context.
\end{itemize}

The remainder of this paper is organized as follows. Section \ref{section_related_works} presents the relevant works. Section \ref{section_system_model} provides a comprehensive overview of the TeleOracle architecture. Our experiments and their corresponding results are detailed in Section \ref{section_simulations}. Finally, section \ref{section_conclusions} concludes the paper with a summary of findings and potential directions for future research.

\section{Related Works}\label{section_related_works}
With growing interest in integrating \acp{LLM} into telecom,
the need for datasets for benchmarking has become increasingly crucial. To this effect, Karim \textit{et al.} introduce a dataset, \textit{SPEC5G}, composed of scraped \ac{3GPP} documents tailored for standards analysis tasks~\cite{karim2023_SPEC5G}. From the created \textit{SPEC5G} dataset, they curate two annotated subsets to evaluate the effectiveness of \textit{SPEC5G} for downstream tasks. To test the effectiveness of their dataset, they pre-train a model on the \textit{SPEC5G} dataset and fine-tune it on one of the curated datasets for downstream tasks. They find that the performance of the pre-trained model surpasses that of the baseline models. 

Nikbakht \textit{et al.}
presents the \textit{TSpec-LLM} dataset~\cite{nikbakht2024_TSPEC}. This dataset covers \ac{3GPP} documents from releases 8 to 19. The authors also develop an automated framework to generate \ac{QnA} pairs. They use a \ac{RAG} model to assess the quality of the dataset and find that supplementing the \ac{LLM} with context from the dataset significantly enhances the model's performance.
Focused on open radio network (O-RAN) specifications, Gajjar \textit{et al.} release \textit{ORAN-Bench-13K}~\cite{gajjar2024_ORAN_Bench_13K}. This benchmark dataset is generated through a three-stage LLM question multiple-generation system from the ORAN specification documents. 
They
also develop ORANSight, a \ac{RAG} framework, and find that this framework enhances the performance of \ac{LLM} on the ORAN \ac{QnA} dataset.

Several works focus on the creation of a \ac{QnA} telecom benchmarking dataset. Holm introduces
\textit{TeleQuAD}, a private benchmark dataset generated for telecom \ac{QnA} tasks~\cite{Holm_2021}. The dataset consists of 4,000 questions and answers generated from randomly sampled sections of telecom documents. Similarly, the \textit{TeleQnA} dataset emerges as a comprehensive contribution to telecom QnA work, encompassing 10,000 entries carefully curated from various sources including 3GPP specifications and research papers~\cite{maatouk2023teleqna}.
They also present a framework for automated generating of \ac{QnA} entries that relies on \acp{LLM} and human validation. 
They evaluate the performance of  GPT-3.5 and GPT-4 on the \textit{TeleQnA} dataset with and without added context in the prompt. Their results confirm the need for a specialized telecom \ac{LM} and the utility of supplementing the prompt with context to better align the models.
Furthermore, 
they evaluate their models against professionals in the field and find that the \ac{LM} perform competitively with humans.


Driven by the \ac{LLM}'s potential, 
several works explore the adaptation of specialized \acp{LLM} for telecom tasks.
Ahmed \textit{et al.} compare re-trained models to fine-tuned models to evaluate the performance of \ac{LLM} in telecommunications tasks~\cite{ahmed2024linguisticintelligencelargelanguage}. On the other hand, Bariah \textit{et al.} focus on fine-tuning \acp{LLM} for text classification purposes~\cite{bariah2023understandingtelecomlanguagelarge}. Another notable example is TeleRoBERTa, developed by Ericsson. TeleRoBERTa is a question-answering LLM model, that is fine-tuned on a substantial corpus of telecom-specific texts, particularly 3GPP specifications~\cite{karapantelakis2024_TeleRoBERTa}. 
These works highlight the potential for these models in the telecom domain.

\ac{RAG} models have also been an area of interest in the field. \textit{Telco-RAG} is a \ac{RAG} model optimized for telecom applications~\cite{bornea2024_telco_RAG}. Their work focuses on conserving random access memory and enhancing the prompt and user's query for better retrieval. They leverage a router to determine the most fitting \ac{3GPP} series to retrieve from and use candidate answer generation to enhance the retrieved chunks.
Similarly, Yilma \textit{et al.} use \ac{RAG} to align the \ac{LLM} to the telecom domain~\cite{yilma2024telecomragtamingtelecomstandards}. Yilma \textit{et al.}, modify the prompt by complementing the user's query with information from their previous queries. Furthermore, they use role-playing prompting techniques to address the user's query in the appropriate depth and complexity.
\begin{figure*}[htbp]
    \centering
    \includegraphics[width=\linewidth]{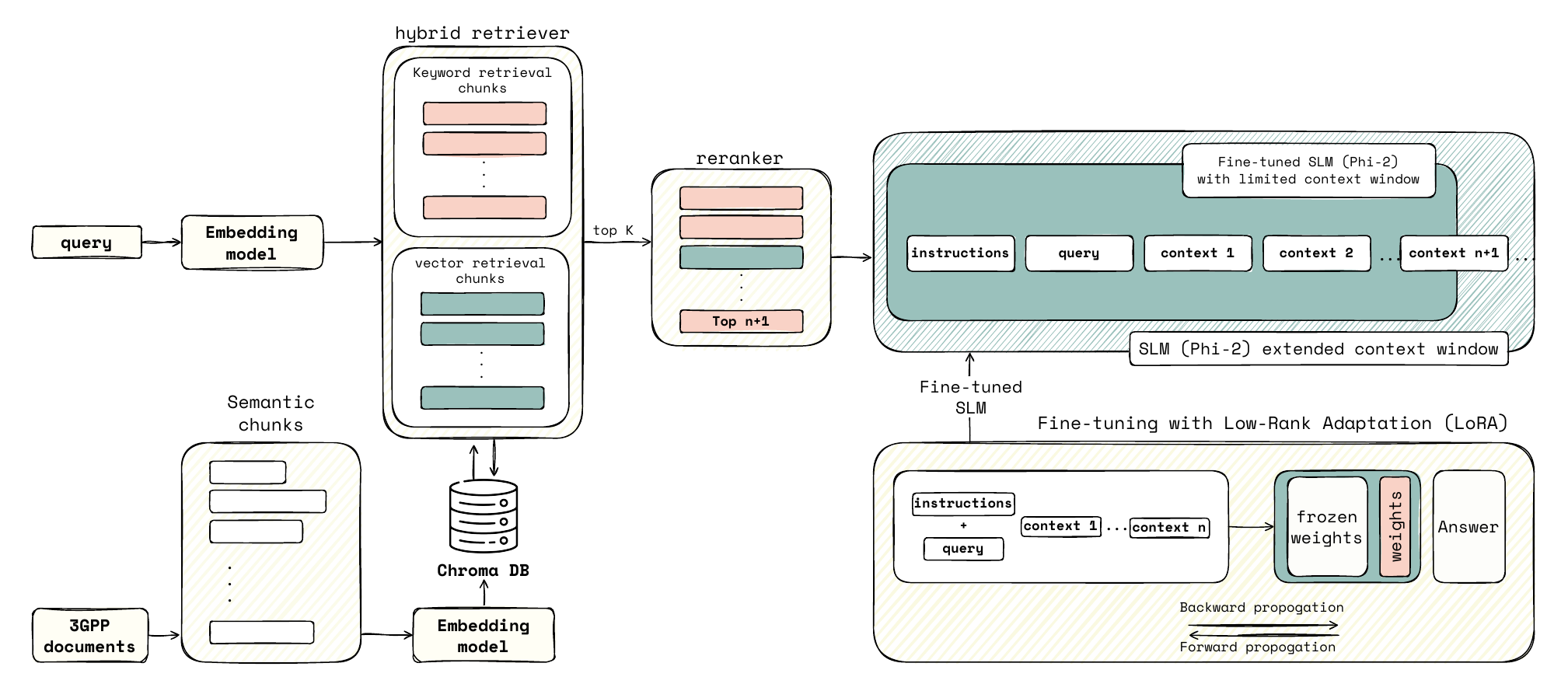}
    \caption{Overview of TeleOracle architecture with semantic chunking, two-stage retrieval process, extended context support at inference time, and fine-tuned Phi-2 SLM integration with \ac{LoRA}.}
    \label{fig:arch}
\end{figure*}

\acp{SLM} have emerged as a notable focus in telecom. 
Piovesan et al. emphasize the necessity of this shift towards smaller models, to overcome edge computing challenges~\cite{piovesan2024_Telecom_LLMs_large}. 
They perform an analysis of the performance of Phi-2 on a downstream telecom \ac{QnA} task and find that \ac{RAG} notably enhances the model's performance underscoring its potential in such tasks.
Recently, Maatouk \textit{et al.} present Tele-LLMs, a collection of telecom-specialized \acp{LLM}~\cite{maatouk2024telellmsseriesspecializedlarge}. They also introduce two telecom datasets, Tele-Data and Tele-Eval, for training and evaluation purposes. Roychowdhury \textit{et al.}
investigate the effectiveness of some of the \ac{RAG} assessment \cite{es2023ragasautomatedevaluationretrieval} metrics on a telecom QA dataset~\cite{roychowdhury2024evaluationragmetricsquestion}. They determine that faithfulness and answer-correctness, which evaluate the relevance of the context and the accuracy of the output respectively, are highly vital in telecom.

While previous works demonstrate the potential of \ac{RAG} \cite{bornea2024_telco_RAG, yilma2024telecomragtamingtelecomstandards} and highlight the importance of SLMs for edge computing \cite{piovesan2024_Telecom_LLMs_large}, TeleOracle introduces new contributions that address key challenges in telecom documentation processing. Unlike existing approaches, we implement semantic chunking and a two-stage retrieval process with a hybrid retriever and re-ranker to handle the contextual complexity of telecom terminology and concepts. Furthermore, by combining SLMs with context window extension, we enable comprehensive responses to complex queries while maintaining the efficiency benefits of smaller models. Our evaluation shows that TeleOracle not only matches larger models' performance but achieves higher faithfulness scores, making it particularly suitable for technical telecom applications.


\section{Proposed architecture}\label{section_system_model}

\subsection{General Overview}
TeleOracle's RAG architecture is composed of four integral components: a semantic document splitter to chunk the documents, an embedding model to encode text to vector representations,  a two-stage retriever for context selection, and an \ac{SLM} generator (Phi-2) to produce the responses. The entire TeleOracle architecture is illustrated in Fig. \ref{fig:arch}.


In the proposed \ac{RAG} framework, the generator, Phi-2, is efficiently fine-tuned using \ac{LoRA}. Furthermore, TeleOracle's setup involves populating a database with semantic chunks of the 3GPP documents (detailed in Sections \ref{sem_chunk} and \ref{embedding}).
After a query is passed to TeleOracle, it is forwarded to the two-stage retriever, consisting of the hybrid retriever and reranker (described in Section \ref{two_stage_ret}). This context is then incorporated into a prompt alongside the user's query, specific instructions for Phi-2, and answer options. To expand the context window and enable the \ac{LM} to handle input sequence beyond the length encountered during fine-tuning, we use SelfExtend, outlined in section \ref{self_extend_section}. In what follows, we elaborate on each of these components.

\subsection{Semantic Chunking Strategy}\label{sem_chunk}

The varying structures of the telecom documents compromise the quality of the database when naive chunking mechanisms are used. This is because the effectiveness of \ac{RAG}'s retrieval mechanism heavily depends on chunking - the process of breaking documents into smaller segments - as it determines not only how information is organized within the embedding system but also how precisely and efficiently relevant context can be retrieved during queries.

\begin{figure}[htbp]
    \centering
    \includegraphics
    [ width=\linewidth]{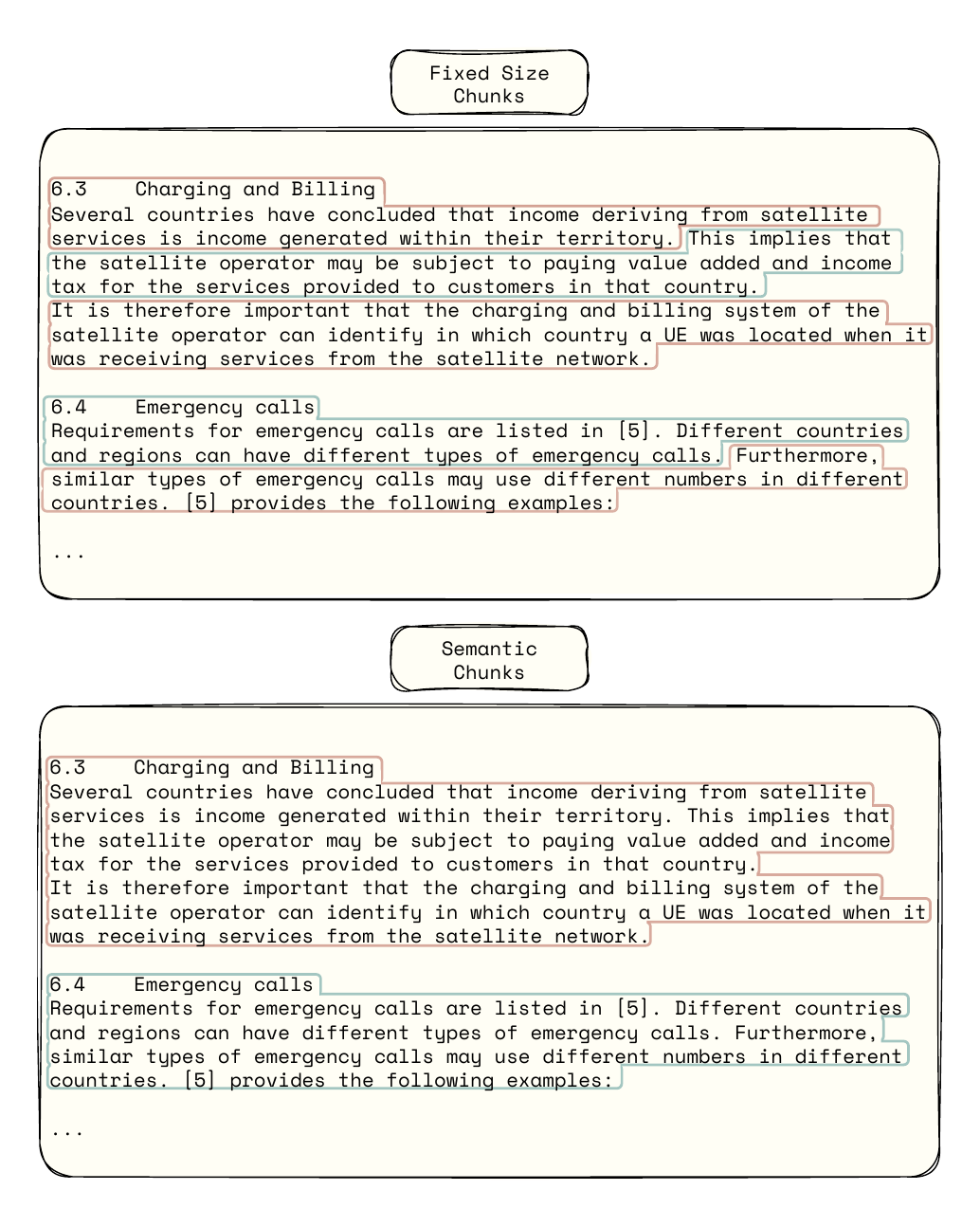}
    \caption{Comparison of fixed-size chunking and semantic chunking applied to an excerpt from a 3GPP document.}
    \label{fig:sem_chunking}
\end{figure}

One of the common chunking methods is fixed-size chunking, which involves 
splitting the chunk based on a fixed chunk length. This method allows the size and number of chunks to be pre-determined to fit the type of documents and the generator's input size limitations.
However, the documents need to be of similar format and a careful study needs to be conducted to determine the best chunk size. In contrast, semantic chunking offers greater flexibility by dynamically determining split points based on a dissimilarity threshold. In this work, we utilize semantic chunking to enhance the relevance and accuracy of the retrieved information. This technique involves splitting the text whenever a set dis-similarity threshold is exceeded. By grouping sentences that are close in the embedding space, we ensure that each chunk is coherent, capturing contextually related information together. Figure \ref{fig:sem_chunking} compares these two approaches, illustrating how semantic chunking better maintains contextual integrity across document segments.

\subsection{Embedding Model and Chroma DB}\label{embedding}
To enable \ac{LM} to process a sequence of text effectively, each element in the sequence must be assigned a numerical representation that captures the semantic meaning and the relationships between elements.
These are known as the vector representations of these elements and are created using the embedding model. 
They are used to evaluate the similarity between the elements. In this paper, we use \textit{bge-small-en-v1.5}~\cite{bge_embedding}, which is optimized for limited resource environments that benefit from a small efficient model. The model is trained using contrastive learning~\cite{chen2020_contrastive}, a technique that entails minimizing the distance between semantically similar pairs while simultaneously maximizing the distance between dissimilar pairs in the embedding space.


Computing these embeddings is time-consuming, thereby making the use of a vector database essential 
for faster performance at runtime. Traditional databases lack optimization for vector-based similarity searches, which involve finding nearest neighbors (i.e., the most similar vectors). This process can be computationally intensive and time-consuming, particularly when handling large datasets. Vector databases, designed specifically for managing high-dimensional embeddings, offer significant performance advantages in these scenarios.
This makes Chroma DB an advantageous choice. Chroma DB is an open-source AI-native vector database specifically designed for managing high-dimensional embeddings. Its dual support for local and remote storage enables organizations to either maintain strict data privacy through local control or leverage cloud-based scalability for edge device deployments. This versatility allows for customized implementations that balance security, performance, and scalability requirements based on specific application needs

\subsection{Two Stage Retrieval}\label{two_stage_ret}

\Ac{RAG}'s performance hinges on its ability to ground generator responses in accurate and relevant context. Our work addresses this through a two-stage retrieval process: a hybrid search retriever paired with a re-ranker.

\subsubsection{Hybrid search retriever}
\begin{figure}[htbp]
    \centering
    \includegraphics[width=\linewidth]{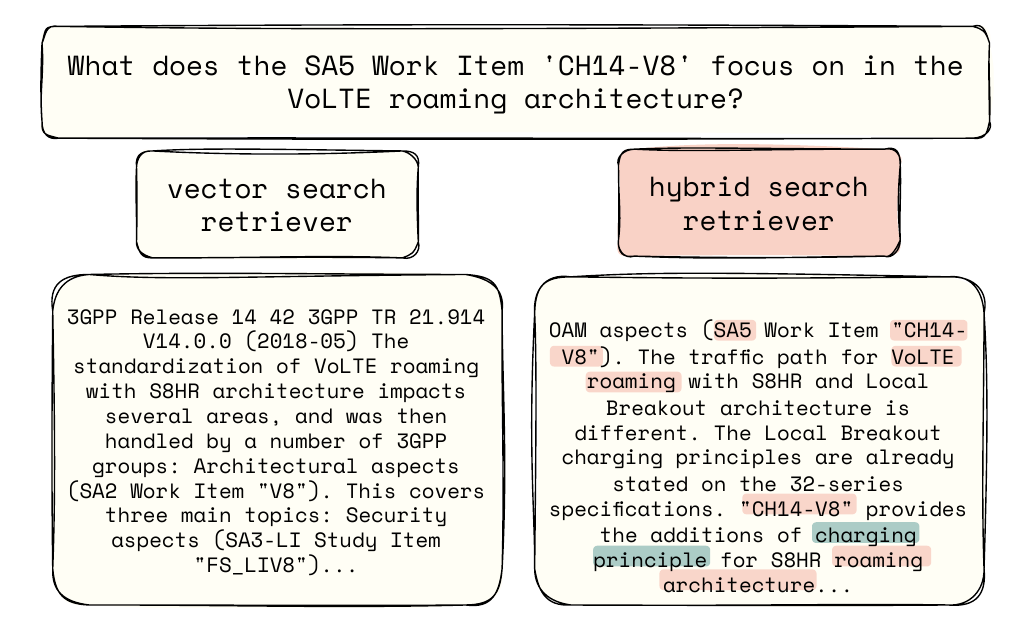}

    \caption{Comparison between vector search and hybrid search retrievers. The retrieved text for the question "What does the SA5 Work Item 'CH14-V8' focus on in the VoLTE roaming architecture?" is shown. The answer to the question is highlighted in blue.}
    \label{fig:hybrid_retreiver}
\end{figure}

We implement a hybrid retriever that leverages the strength of both semantic and keyword-based search to enhance context retrieval. 

Keyword search retrievers are used to find chunks based on specific words or phrases.
Having a keyword-based retriever is essential for telecom applications because of the presence of telecom-specific terms, such as protocol or document names that must be accurately identified and retrieved. For instance, in Fig.~\ref{fig:hybrid_retreiver}, the question references several key terms such as 'SA5'  (Service and System Aspects 5), 'CH14-V8' (an identifier for a specific work item within the SA5), VoLTE (Voice over LTE, and roaming architecture. These terms are crucial for understanding the context and accurately addressing the query.

In this work, we use BM25~\cite{askari2023injecting} as our keyword search retriever, which scores queries as follows,
\begin{equation}
\begin{aligned}
BM25\_score = & \frac{f(q,D) \times (k+1)}{f(q,D) + k \times \left(1 - b + b \times \frac{dl}{adl}\right)} \\
& \times \log\left(\frac{N - N(q) + 0.5}{N(q) + 0.5} + 1\right),
\end{aligned}
\end{equation}
%
%
where $f(q,D)$ is the frequency of the query term $q$ in document $D$, $dl$ is the length of document $D$, $adl$ is the average document length across the collection,  $N$ is the total number of documents in the collection, and $N(q)$ is the number of documents containing the query term q. 
and $log(\frac{N}{N(q)})$ is the inverse document frequency. 
The parameters $k$ and $b$ are used to fine-tune the formula, where $k$ controls term frequency saturation and $b$ adjusts for document length normalization.

BM25 is needed in our case to retrieve telecom-specific terminology because it effectively ranks documents based on the presence and frequency of these specific terms. By prioritizing documents that contain these critical keywords, it ensures that the most relevant information is retrieved. This retriever improves on an existing similarity search method, namely \ac{TF-IDF}.  \Ac{TF-IDF} calculates the relevance of a term based on its 
term frequency and the inverse document frequency, which is the measure of how rare or common a term is across the entire corpus. 
This allows it to score terms that occur often in all documents, such as 'The,' lower compared to terms that occur often only in a few documents such as 'NWDAF.'

\ac{TF-IDF} identifies important terms while differentiating between common and rare terms. However, there are more factors to consider when deciding on the relevance of the document. BM25 builds on \ac{TF-IDF} and addresses those factors. For instance, the frequency of a term $TF$ is not directly proportional to its relevance. In other words, if the term 'NWDAF' occurs 40 times in one document and 20 times in another, the document with 40 occurrences is not necessarily twice as relevant as the one with 20 occurrences. Therefore, the term $k$ is added to moderate the influence of the term frequency on the score.
Furthermore, BM25 considers the influence of the length of the document on the significance of term frequency  $TF$. A term appearing twice in a short document can be more significant than a term that appears twice in a much longer document. BM25 introduces a new term that penalizes longer documents $\frac{dl}{adl}$. These additional components render the BM25 a better fit for this work than the \ac{TF-IDF}. 

Vector search relies on identifying the semantic similarity between vectors, measured by the relative distances between them. The vector search retriever first vectorizes the query using the embedding model, converting it into a high-dimensional embedding. This vector captures the query's meaning. The retriever then calculates the cosine similarity between this query's vector and the existing vectors stored in the database. Cosine similarity, expressed as the dot product of two vectors over their magnitude, presents as a low-cost operation that captures the similarity between vectors.
Vector search excels at understanding the context and nuances of complex queries, which is crucial for retrieving context that adequately addresses telecom-related questions.

After a list of context chunks is retrieved by the semantic and keyword-based search, a merged list is created by taking the union of both lists of retrieved chunks~\cite{Bruch2022_Sebastian_hybridRetr}. Hybrid search retrievers capture intricate details and words while also understanding the overall context. 
Fig~\ref{fig:hybrid_retreiver} demonstrates an example from our experiments where the hybrid retriever provides a more relevant output compared to only using a vector retriever. 



\subsubsection{Re-ranking}
\begin{figure}[htbp]
    \centering
    \includegraphics
    [ width=\linewidth]{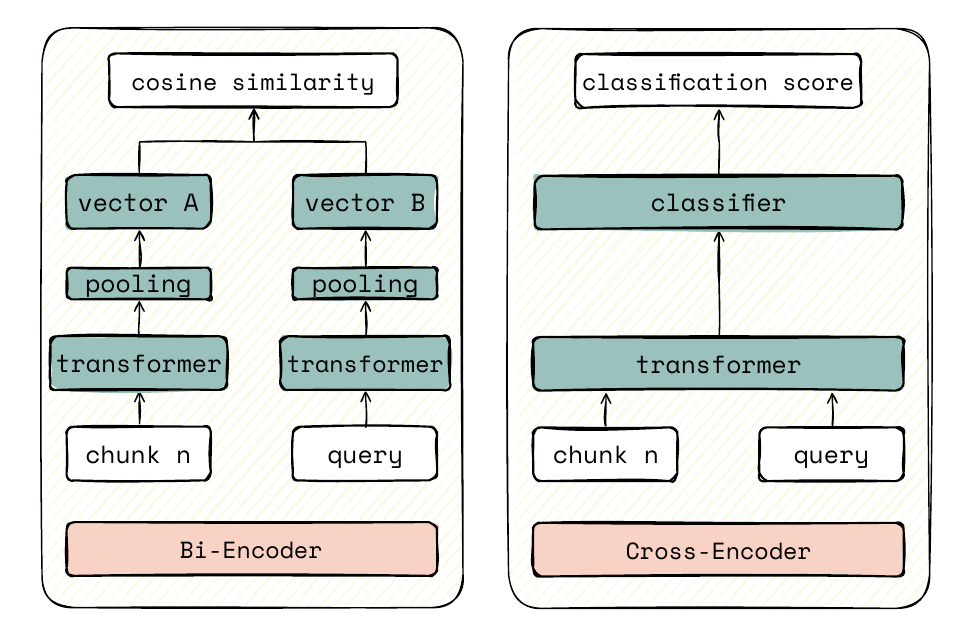}
    \caption{Bi-Encoder vs. Cross-Encoder architectures. In the Bi-Encoder, query and chunk are encoded separately, and cosine similarity is used to compare embeddings. The Cross-Encoder jointly encodes the query and chunk, producing a classification score.}
    \label{fig:cross_encoder}
\end{figure}
The first retrieval stage, the hybrid retriever, is computationally inexpensive but does not capture the fine-grained contextual relationships between the query and documents.
Relevant context concealed within the middle of the contexts list tends to be overlooked by the \ac{LLM}~\cite{liu2023lostmiddlelanguagemodels}.
This highlights the need to refine the list of contexts. To achieve this we incorporate a cross-encoder reranker into our framework~\cite{huggingface_msmarco_minilm_l_6_v2}. Cross encoders, illustrated in Fig. \ref{fig:cross_encoder}, evaluate the similarity between each query and document pair and then output a classification score between between 0 and 1, 
where 1 signifies relevance and 0 indicates irrelevance.
This allows cross-encoders to capture nuances between context chunks the query, however, the pairwise calculations make this a time-consuming and computationally expensive process.

On the other hand, bi-encoders, used in stage one, encode all the documents in advance before performing a search, enabling faster and more efficient retrieval. However, they typically underperform compared to cross-encoders in terms of accuracy, as they may fail to capture complex interactions between the query and context chucks.
This further highlights the trade-off between efficient performance and accuracy and demonstrates that a two-stage retrieval, such as the one employed in this framework, is essential to maintain an acceptable inference time, while providing optimal accuracy.


\begin{figure}[htbp]
    \centering
    \includegraphics
    [ width=\linewidth]{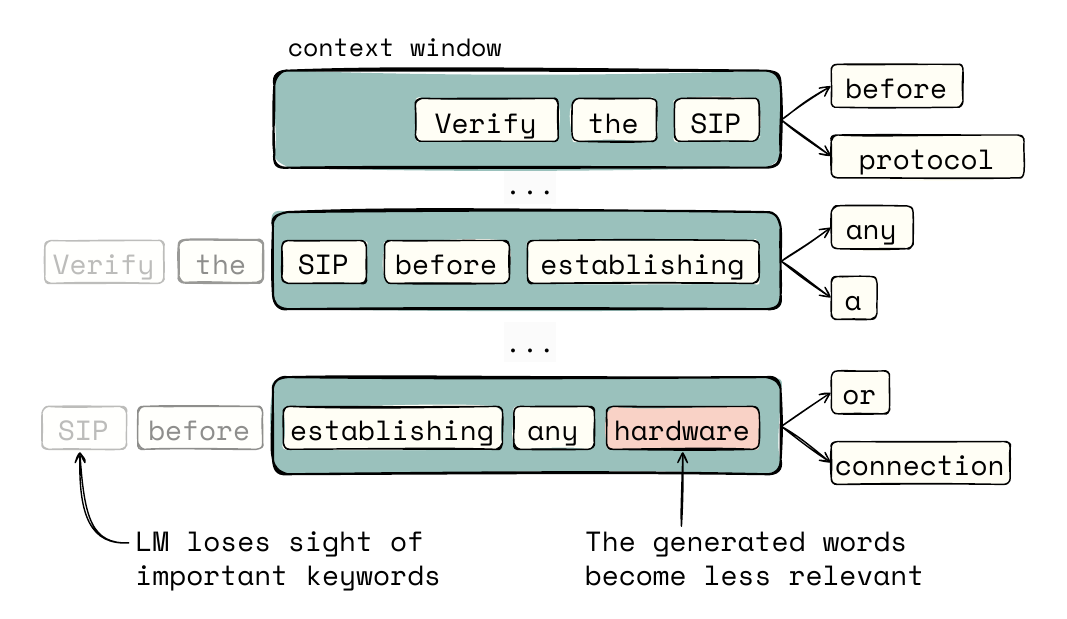}
    \caption{ Example of how language models lose important context when input sequences exceed their fixed context window. As the sequence length grows, the model forgets key tokens like "SIP" and generates less relevant outputs, such as "hardware," due to its inability to retain words beyond the context window.}
    \label{fig:L_seq_eg}
\end{figure}

\begin{figure}[htbp]
    \centering
    \includegraphics
    [width=\linewidth]{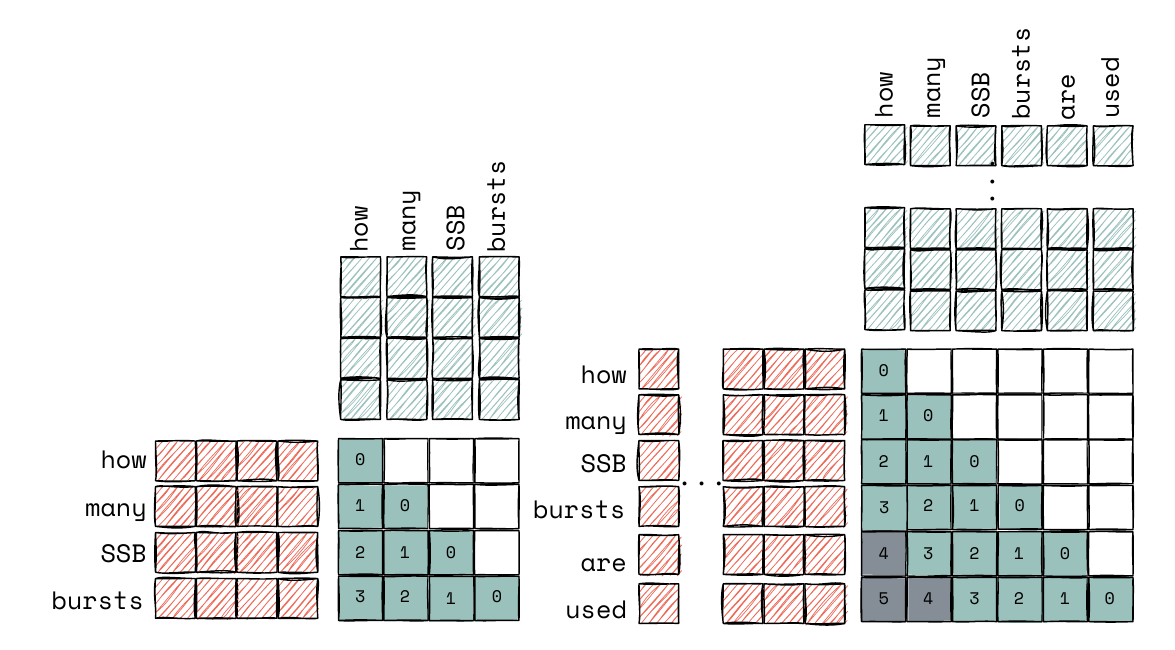}
    \caption{Illustration of the positional Out-of-Distribution (OOD) issue. The figure demonstrates the relative distances when the input sequence length is within the pre-trained model’s context window (left), and how unseen relative distances
    arise when the input sequence exceeds the model’s context window (right).}
    
    \label{fig:attn_long}
\end{figure}



\subsection{Extending the Context Window with SelfExtend}\label{self_extend_section}
During training, \acp{LM} are exposed to a fixed context window length, i.e., an input sequence of a fixed length. If the context window's length during inference exceeds what the \ac{LM} was trained on, it behaves unpredictably and produces gibberish outputs. This occurs because \acp{LM} cannot retain words that extend beyond the context window, this is illustrated in Fig.~\ref{fig:L_seq_eg}.

Though this inability to generalize may not present as an issue for larger propriety models with large context windows, it becomes particularly problematic for smaller open-source language models. These models are trained on input small sequences due to limitations in computational resources and the available training data. Addressing this limitation is critical as chunking techniques such as semantic chunking, and future improvements such as adding tables to the context, the context window of the \acp{LM} will need to be expanded. Furthermore, a bigger number of chunks might be retrieved to adequately answer broad questions
For instance, \textit{"From a UE perspective, what does the UE need to be aware of?"} and \textit{"How are messages routed in the 5G system?"} are inherently broad questions. They could touch upon numerous aspects such as network registration, protocol compliance, and so on. To adequately respond the generator needs to be supplemented with a substantial number of chunks.

Furthermore, in our tests, we exclude the \ac{MCQ} options from the query during retrieval to simulate a more realistic user interaction when posing a question. This approach ensures that the \ac{LM} does not rely on pre-provided answer choices, thereby preserving the complexity of the task. Certain questions in the dataset, such as \textit{"From a BS perspective, which statement is true?"}
and \textit{"Which of the following is NOT a function of XnAP?"}, benefit from the inclusion of options as they help narrow down the retrieved text. Without the provided options, a more extensive list of retrieved-context would be necessary to answer these types of questions accurately.

It is also wrong to assume that all users of the telecom oracle are proficient in prompting \ac{LM}. For instance, inexperienced users might use queries that might be vague, ambiguous, or lack the specificity needed to extract the most relevant information. Providing the \ac{LM} with a wide array of context could allow it to sufficiently address the user's query.

The issue of input sequences longer than the pre-trained model's context windows has been addressed in several works.
Most proposed techniques involve fine-tuning in a longer context \cite{PI, CLEX, YARN}. However, fine-tuning requires extensive resources. the work in \cite{self_extend} proposes SelfExtend, a method to extend \acp{LM} context window without requiring fine-tuning. SelfExtend alters the way in which input is passed to the self-attention mechanism by modifying the positional encoding. Self-attention is a foundational concept in transformers that involves processing tokens in the input sequence in parallel. It requires position encoding to be applied to each token to preserve information about the word order.
In other words, the tokens are assigned values that represent their position, providing useful information to the \ac{LM} about the relationship between tokens in the sequence. The authors in \cite{self_extend} show that altering this encoding can unlock the \ac{LM}'s ability to handle longer input sequences.
SelfExtend can be applied to models that use \ac{RoPE}, one of the many to encode positional information for each token~\cite{su2023roformerenhancedtransformerrotary}. Other positional encoding methods such as absolute positional encoding either lack the ability to generalize to unseen positions, or struggle to handle long sequences. \ac{RoPE} uses a rotation matrix where the token's absolute positional encoding is rotated depending on its position in a sentence.
Though \ac{RoPE} is theoretically capable of handling longer sequences, it faces challenges in practice, namely, its inability to distinguish between distant values as the rotational angles become too large, leading to a loss of positional accuracy. 
When the model is exposed to sequences longer than it encountered during training, these tokens are assigned positions that fall outside the distribution of the training data as illustrated in Fig.~\ref{fig:attn_long}. 
This suggests that 
the models struggle with longer input sequences not because of an inherent lack of ability to handle long input sequences, but rather due to the model overfitting on the positional encoding it was exposed to during fine-tuning. SelfExtend 
builds on the idea that relative positions do not have to be unique, and uses the floor operation resulting in position indices up to the maximum position index seen during training. This approach, where a few tokens are mapped to the same position embedding, is referred to as grouped attention. The relative position of the token is more critical to the understanding of the sequence of text than the exact positions~\cite{self_extend}. Additionally, the order in which words appear in shorter contexts tends to have limited valid orders. 
Therefore, for larger input sequences, the grouping does not interfere with the method's performance.

\begin{figure}[htbp]
    \centering
    \includegraphics
    [width=\linewidth]{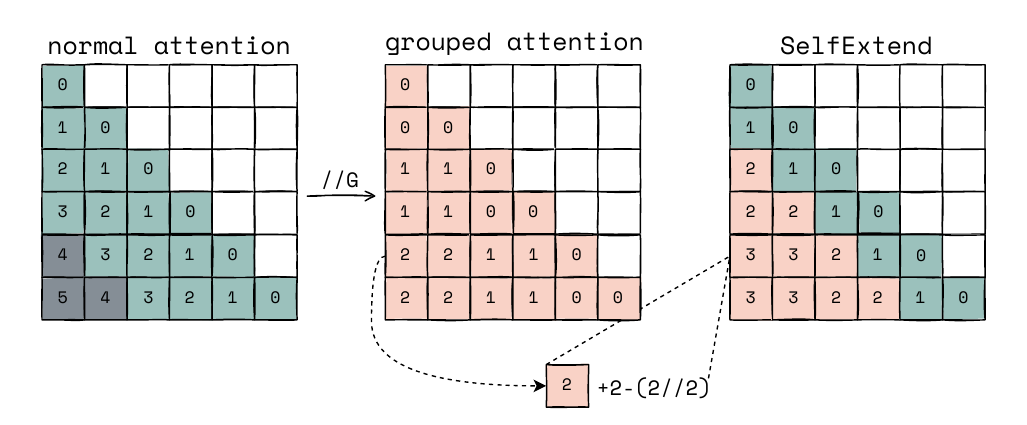}
    \caption{Illustration of how SelfExtend handles input sequences longer than the pre-trained model's context window size. In the figure, the model's context window size, $L$, is 4. The neighbor window, $w_n$ is set to 2, and the group size,$G_s$, is set to 2. Relative positions beyond $w_n$ are calculated using $w_n-\floor*{\frac{w_n}{G_s}}$.} 
    \label{fig:self_extend}
\end{figure}

The former method alone hinders the model's performance and reduces the coherence of the generated text. This is because for immediate neighboring tokens, the exact position matters. For instance, "\textit{Only one UE connects to the base station.}" is very different from "\textit{One UE only connects to the base station}", making it essential to maintain the exact values when looking at immediate neighboring tokens. Therefore, SelfExtend also applies the standard attention to the tokens closest in proximity to the target token. They refer to this as the neighbor attention. Fig.~\ref{fig:self_extend} illustrates the steps taken by SelfExtend to alter the positional encoding. 

Using SelfExtend, we extend the context window of the pre-trained Phi-2's from 2048 tokens to a maximum of 8192. This extension enables better prompting and facilitates the incorporation of tables, a common structured data type in technical documents, and other large data into the model's input, further enhancing its performance.

\subsection{The Generator: Fine-tuned Phi-2 with Multiple Contexts}

One of the contributors to the \ac{LM}'s inferior performance in specialized technical domains, such as telecom, is the
prevalence of technical terms and concepts primarily appearing in white papers and specification documents. To circumvent this, we fine-tune Phi-2, ensuring 
that the model is adept at adequately leveraging the given context to answer the query and to familiarize it with common terminology terms.
To make the fine-tuning process more accessible and reduce the computational and memory demands we utilize gradient accumulation and \ac{LoRA}~\cite{Edward_2021_LoRA}.

Gradient accumulation enables the model to overcome the memory constraint of training on larger batch sizes by accumulating gradients from multiple smaller batches before updating the weights. The issue with simply using smaller batch sizes is that it could result in noisy updates and affect convergence. Gradient accumulation simulates larger batch sizes, such as 256 which may not be accommodated by the hardware, by creating 8 batches of size of size 32 each and updating the model's weights with the accumulated gradient.

Though pre-trained \acp{LM} might have a large number of parameters, it has been demonstrated that the essential information is concentrated in a subset of these parameters, which is referred to as the model's 'intrinsic dimension'~\cite{Aghajanyan2020}. The low intrinsic dimension of models enables the use of techniques, such as \ac{LoRA} to focus on and efficiently update this core subset during fine-tuning.
During fine-tuning, weight updates $W_{new}$ are represented as 
\begin{equation}
W_{new} = W_0 + \Delta W,
\label{eq:weight_update}
\end{equation}
where $W_0 \in \mathcal{R}^{d\times k}$ are the initial pre-trained weights, and $\Delta W \in \mathcal{R}^{d\times k}$ represents the change in weights. Computing $\Delta W$ is resource-intensive due to the large scale of the matrices involved and the intensive matrix multiplications required for each weight update. In \ac{LoRA}, trainable parameters $\Delta W$ are expressed as a product of two low-rank matrices, $B\times A$, where $B\in \mathcal{R}^{d\times r}$  and $A\in \mathcal{R}^{r\times k}$, with rank $r \ll \min(d,k)$. Thus, with \ac{LoRA}, weight updates $W_{new}$ are computed as 
\begin{equation}
W_{new} = W_0 + (\frac{\alpha}{r}) B \times A.
\label{eq:weight_update}
\end{equation}
Parameter $\alpha$ serves as a scaling factor, reflecting the impact of the 
weight updates on the initial pre-trained weights. Therefore, due to this matrix decomposition, only $d \times r + r \times k $ parameters require updating.
 

Though other techniques, such as using adapter layers \cite{Adapter}, also seek to reduce the number of parameters trained during fine-tuning and make the process more efficient, They often introduce additional latency at inference time. This is not the case with \ac{LoRA}. In \ac{LoRA}, the optimized weights are merged with the frozen weights of the base model during deployment, ensuring efficient parameter utilization without additional latency. Real-time processing is essential for telecom applications given the time-sensitive nature of tasks such as network monitoring, fraud detection, and customer support, for which this work is a precursor.

\Ac{QLoRA} is variation of \ac{LoRA} in which the model is quantized to 4-bit precision\cite{Qlora}. \Ac{QLoRA} effectively reduces the model's size, increasing computational efficiency and maintaining low resource requirements, without compromising performance. This is particularly significant because maintaining reliable accuracy is essential for telecom applications.
%
%
\ac{LoRA} and its variation \ac{QLoRA}, offer enhanced improved memory efficiency both in the training and in storing the fine-tuned models. \ac{LoRA} stores the base model separately from the fine-tuned weights, significantly reducing the memory footprint of the fine-tuned model. This decoupling of the base model from task-specific weights enables efficient storage and streamlined management.






\begin{figure}[htbp]
    \centering
    \includegraphics
    [width=0.8\linewidth]
    {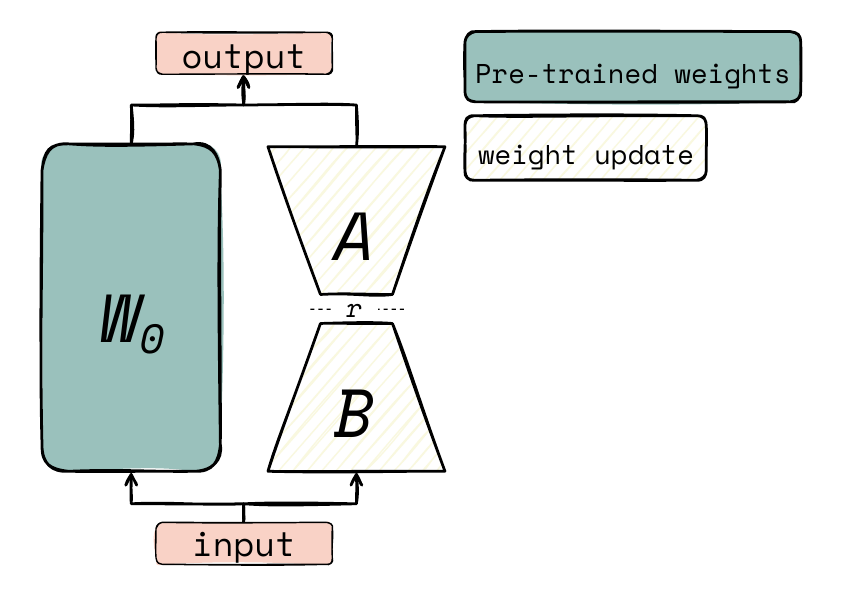}
    \caption{Schematic illustration of the Low-Rank Adaptation (LoRA) technique for efficient fine-tuning of neural networks with low-rank matrices (e.g., \acp{LM}).}
    \label{fig:LoRA}
\end{figure}

\subsection{Prompt Engineering}
\begin{figure}[htbp]
    \centering

    \includegraphics[width=\columnwidth]{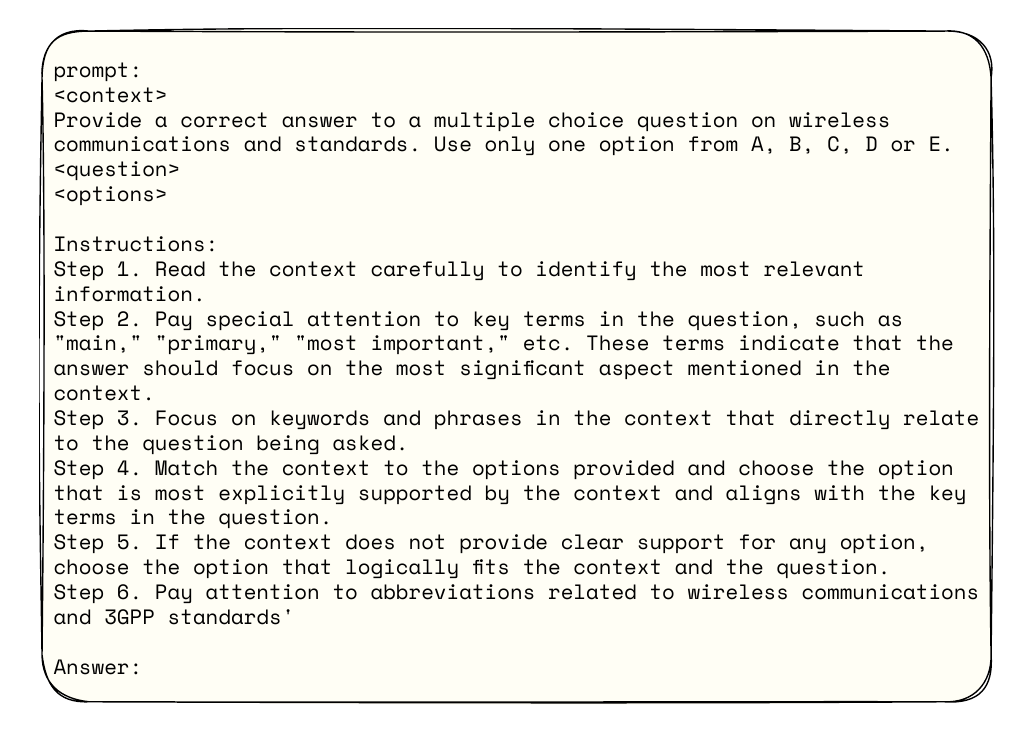}
    \caption{Prompt structure that includes retrieved context and instructions.}
    \label{fig:prompt}
\end{figure}

To fully leverage the capabilities of \ac{SLM}, effective prompting must be implemented. In this work, the query is placed in the prompt with a set of instructions and the retrieved relevant context, as seen in Fig. \ref{fig:prompt}. The instructions serve to guide the \ac{LM} through the steps that it should take before answering the question. They bring attention to critical key terms such as ``main`` and ``primary`` and telecom-specific language which can be instrumental in correctly interpreting the nuances of the question and identifying the most relevant context.







The \ac{LM} is also guided towards adequately relying on the provided context, ensuring that the model's prior knowledge is less likely to interfere with the language model's response. Lastly, the instructions are also a means to ensure that the outputs of Phi-2 are uniform to ensure the answers can be extracted from the output.

\section{Experimental results}\label{section_simulations}
\subsection{Settings}
The \ac{RAG} framework leverages a database containing approximately 550 semantically chunked 3GPP documents, encompassing specifications up to Release 18. For semantic chunking, a breakpoint percentile threshold of 90, and a buffer size of 3, specifying that three sentences would be grouped as a unit when comparing similarity, and that a breakpoint is created when cosine dissimilarity of 90 is reached.

To finetune the generator, we utilize the TeleQnA dataset~\cite{maatouk2023teleqna}, which comprises 10,000 multiple-choice questions categorized by topic. Each question entry has up to five options, a single correct answer, and an accompanying justification for the solution. The testing dataset consists of a set of 2000 questions~\cite{Zindi}, each entry consisting of a question with and up to five answer options. The type of questions include true/false statements, inquiries about specifications and definitions, and questions regarding the purpose and role of entities.



For fine-tuning, the following hyperparameters are set: learning rate of $10^{-4}$, batch size of 32, dropout rate of 0.05, weight decay of 0.01. The rank and alpha for \ac{LoRA} are 32 and 64, respectively. 
The hybrid search retriever is configured to retrieve the top 150 relevant chunks and the reranker is set to return the top 15 chunks from the 150 retrieved chunks. To account for the chunk size variance introduced when using a semantic chunker, we employ a second query engine that retrieves extra chunks from a database with fixed-size chunks until a set minimum token size is reached. We perform this to ensure that the context window is fully utilized.  
The aforementioned hyper-parameters are not necessarily optimized, but rather set based on qualitative judgment and limited exploration.
For reproducibility and reuse, our source code is made publicly available
\footnote{source code available at \href{https://github.com/Nouf-Alabbasi/oKUmura_AI_Telecom_challenge}{https://github.com/Nouf-Alabbasi/oKUmura\_AI\_Telecom\_challenge}}.




\subsection{Results and Analysis}
\begin{table}[htbp]
    \centering
    \caption{Accuracy comparison of our fine-tuned Phi-2 based RAG model against baseline models on telecom standards questions, both with and without retrieved context.}
    \label{tab:models_eval}
    \includegraphics{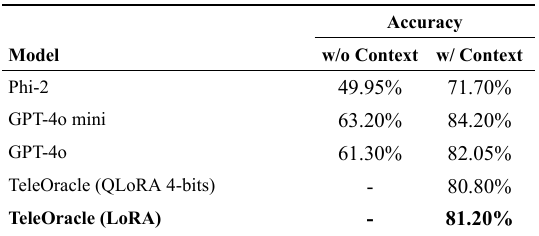}
\end{table}

In table \ref{tab:models_eval}, we benchmark our model against base Phi-2 (2.7B), GPT-4o mini , and GPT4o.
We find that the base Phi-2 model considerably underperforms.
Similarly, we find that leveraging retrieved context noticeably enhances all the benchmarked models' performance, indicating the efficacy of using \ac{RAG} in this context.

\begin{table}[htbp]
    \centering
    \caption{Faithfulness scores of TeleOracle, GPT-4o mini, and GPT-4o. TeleOracle achieves the highest score of 78.80\%, indicating stronger alignment between its responses and the retrieved context compared to the other models.}
    \label{tab:faith}
    \includegraphics{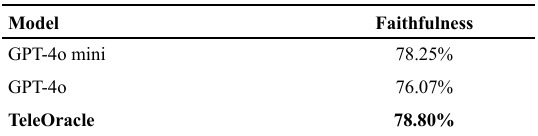}
\end{table}

Notably, GPT-4o achieves only 82.05\% accuracy, performing slightly worse than its smaller counterpart. This lower performance can be attributed to its highly generalized nature, which leads to misinterpretation of specialized telecommunications terminology and concepts.
For instance, The output GPT-4 provides for the following \ac{MCQ} question indicates a level of knowledge interference:

\textit{How can the impact of noise in MPAC on FR1 MIMO OTA performance be corrected?}

\textit{1. By decreasing the attenuation value}

\textit{2. By increasing the attenuation value}

\textit{3. By replacing the amplifiers}

\textit{4. By adjusting the amplifier output power}

Both TeleOracle and GPT-4o mini pick the correct answer choice, namely option 2. However, GPT-4o choses option 4 which states that the noise impact in MPAC (Multi-Probe Anechoic Chamber) on FR1 MIMO OTA (Over-The-Air) performance can be corrected '\textit{by adjusting the amplifier output power}'. This indicates that GPT-4o is relying on its acquired general knowledge and is conflating engineering principles with telecom-specific principles. In general engineering principles, increasing signal power is a common method to improve performance in noisy environments. However, in the specific context of MPAC testing for FR1 MIMO OTA performance, increasing the attenuation value is the correct method to mitigate the impact of noise.

We use the Faithfulness metric to quantify this effect~\cite{es2023ragasautomatedevaluationretrieval}. This metric measures the extent to which the model's response is informed by the retrieved context. Higher faithfulness indicates stronger alignment between the context and the generated output. The results of this evaluation are shown in Table~\ref{tab:faith}, where GPT-4o performs notably worse than TeleOracle and GPT-4o mini, reflecting its reduced ability to utilize the retrieved context effectively.




Furthermore, the results in Wu \textit{et al.} work suggest that there is a relation between the model's confidence in its prior knowledge and adherence to the retrieved context~\cite{internal_prior_and_external_evidence}. This further reinforces our hypothesis that aligning the model through fine-tuning is crucial for improving its accuracy and reducing potential conflations.
Additionally,  despite clear instructions provided, GPT4-o does not consistently adhere to the required output format, leading to the predicted answer option not being extracted properly. It is likely that larger models, such as GPT-4o, overfit on generating free-form, coherent text, making them less responsive to structured output requirements. This further underscores the importance of fine-tuning the model to overcome these challenges, given that adhering to the required format holds extreme significance in certain downstream applications in telecom. 

We also note that the model fine-tuned using \ac{QLoRA} achieves comparable performance to that of the model fine-tuned using \ac{LoRA} with accuracies 80.80\% and 81.20\% respectively. Quantizing a model during fine-tuning can save memory and computational resources, but it can also introduce small inaccuracies in the model's representations, leading to a minor drop in performance. The fact that this degradation is only slight indicates that \ac{QLoRA} and \ac{LoRA} are robust to quantization, maintaining much of their effectiveness even in a more compressed form. From these results, it is evident that fine-tuning the \ac{SLM} and utilizing our \ac{RAG} framework are effective techniques to align Phi-2 to the telecom domain.

\begin{table}[htbp]
    \centering
    \caption{Performance comparison of various configurations of the fine-tuned Phi-2 model with RAG and additional components on telecom specification \ac{QnA}. The table uses the following acronyms: SE for SelfExtend, RR for Rerank, SC for Semantic Chunking, and MC for Multiple Context.}
    \label{tab:components}
    \includegraphics{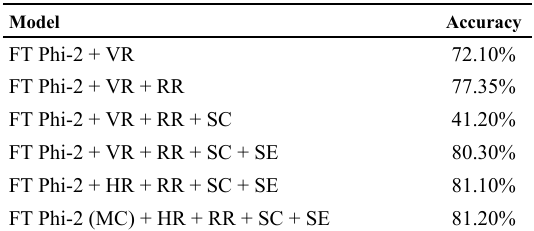}
\end{table}

To further analyze the results and the impact of each involved component in the architecture, we perform an ablation study. This analysis reveals that Re-ranking, with an accuracy of 77\%, 
contributes significantly to the overall performance improvement.
The findings underscore the importance of a two-stage retrieval process, which effectively captures the intricate relation between the query and context chunks while maintaining an acceptable retrieval speed.
While the addition of the hybrid search retriever has a marginal effect on the base model's accuracy (only improving accuracy by around 1\%), it plays a crucial role by capturing important telecom-specific keywords within the retrieved context chunks that the vector retriever alone might overlook, as demonstrated in Fig. \ref{fig:hybrid_retreiver}.

It is significant to point out that the impact of selfExtend becomes more pronounced when semantic chunking is used resulting in an 8\% increase compared to the base model, raising the accuracy to 80.30\%. 
 Semantic chunking produces more cohesive chunks, although they tend to be longer, as shown in Fig. \ref{fig:sem_chunking}.
%
SelfExtend extends the context window to allow the incorporation of these chunks into the prompt while ensuring that critical information is retained.
We also observe the effect of fine-tuning with multiple retrieved contexts in the prompt. This enhancement, illustrated by the slight improvement in accuracy, can be attributed to the model's improved ability to leverage nuanced ideas from the context and draw more accurate inferences.

\section{Conclusions And Future Works}\label{section_conclusions}

The complex nature of telecom documentation necessitates specialized language models that can handle the domain's nuances. In this work, we present TeleOracle, an advanced \ac{RAG} framework that leverages \ac{SLM}s for telecommunications tasks. By integrating \ac{LoRA}, context window extension, semantic chunking, and a two-stage retrieval process, our system efficiently processes complex documentation while maintaining computational efficiency. Our experimental results demonstrate that TeleOracle outperforms baseline models in both accuracy and faithfulness, demonstrating the viability of specialized \acp{SLM} for telecommunications applications and establishing a foundation for efficient language-based solutions in the field.

A significant limitation of \ac{RAG} models is their inability to effectively incorporate tables, figures, network graphs, and communication-specific features such as channel information~\cite{meng2023netgptgenerativepretrainedtransformer}. These data types contain valuable information that can enhance the \ac{LM}'s responses. Future work can focus on integrating these elements into the retrieved context. Beyond data integration challenges, certain questions and tasks require careful reasoning and planning. For example, queries like \textit{'How can the UPF assist RAN with identifying the end of DL traffic bursts?'} demand complex analysis that may exceed the capabilities of one-shot prompting. This limitation highlights the need to explore advanced reasoning techniques and domain-specific approaches.



\bibliographystyle{IEEEtran}
\bibliography{ref}

\end{document}